\documentclass{article}

\usepackage{kafkanotes}
\usepackage{bm}
\usepackage{cleveref}
\usepackage{natbib}

\title{An Introduction to Transformers}
\author{Richard E.~Turner\\ Department of Engineering, University of Cambridge, UK\\ Microsoft Research, Cambridge, UK\\
\tt{ret26@cam.ac.uk}}

\begin{document}

\begin{titlepage}
\thispagestyle{empty}
\maketitle

%Abstract
\begin{abstract}
The transformer is a neural network component that can be used to learn useful representations of sequences or sets of data-points \citep{NIPS2017_3f5ee243}. The transformer has driven recent advances in natural language processing  \citep{devlin-etal-2019-bert}, computer vision \citep{vision-transformer}, and spatio-temporal modelling \citep{bi2022panguweather}. There are many introductions to transformers, but most do not contain precise mathematical descriptions of the architecture and the intuitions behind the design choices are often also missing.\footnote{See \cite{phuong2022formal} for an exception to this.} Moreover, as research takes a winding path, the explanations for the components of the transformer can be idiosyncratic. In this note we aim for a mathematically precise, intuitive, and clean description of the transformer architecture. We will not discuss training as this is rather standard. We  assume that the reader is familiar with fundamental topics in machine learning including multi-layer perceptrons, linear transformations, softmax functions and basic probability.

\end{abstract}

%\tableofcontents
\end{titlepage}

%Geometri untuk halaman konten
\newgeometry{top=20mm,bottom=25mm,right=80mm,left=20mm}

%================KONTEN DIMULAI DISINI================%

\section{Preliminaries}

Let's start by talking about the form of the data that is input into a transformer, the goal of the transformer, and the form of its output.

\subsection{Input data format: sets or sequences of tokens}

\begin{marginfigure}%
   \includegraphics[width=0.8\linewidth]{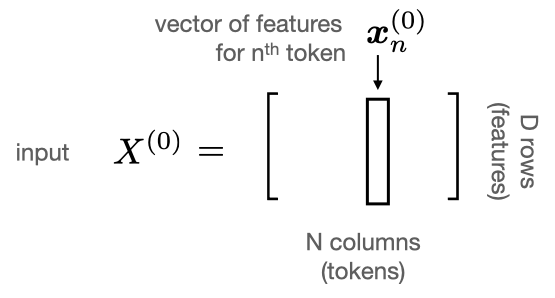}
   \caption{The input to a transformer is $N$ vectors $\bm{x}_n^{(0)}$ which are each $D$ dimensional. These can be collected together into an array $X^{(0)}$.}
   \label{fig:input}
 \end{marginfigure}
 
\begin{marginfigure}%
   \includegraphics[width=0.8\linewidth]{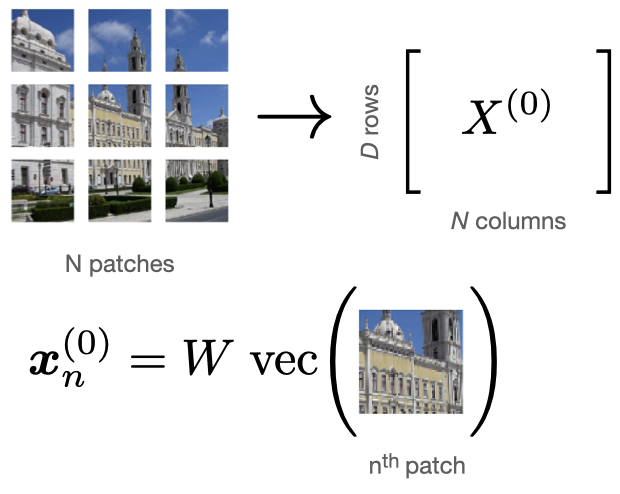}
   \caption{Encoding an image: an example \citep{vision-transformer}. An image is split into $N$ patches. Each patch is reshaped into a vector by the $\text{vec}$ operator. This vector is acted upon by a matrix $W$ which maps the patch to a $D$ dimensional vector $\bm{x}^{(0)}_n$. These vectors are collected together into the input $X^{(0)}$. The matrix $W$ can be learned with the rest of the transformer's parameters.}
   \label{fig:image-embedding}
 \end{marginfigure}

In order to apply a transformer, data must be converted into a set or sequence\sidenote{\footnotesize Strictly speaking, the collection of tokens does not need to have an order and the transformer can handle them as a set (where order does not matter), rather than a sequence. See section \ref{sec:pos-ecoding}.} 
of $N$ tokens $\bm{x}^{(0)}_n$ of dimension $D$ (see figure \ref{fig:input}). The tokens can be collected into a matrix $X^{(0)}$ which is $D \times N$.\sidenote{\footnotesize Note that much of the literature uses the transposed notation whereby the data matrix is $N \times D$, but I want sequences to run across the page and features down it in the schematics (a convention I use in other lecture notes).} To give two concrete examples 
\begin{enumerate}
\item a passage of text can be broken up into a sequence of  words or sub-words, with each word being represented by a single unique vector, 
\item an image can be broken up into a set of patches and each patch can be mapped into a vector.
\end{enumerate}
The embeddings can be fixed or they can be learned with the rest of the parameters of the model e.g.~the vectors representing words can be optimised or a learned linear transform can be used to embed image patches (see figure \ref{fig:image-embedding}). 

A sequence of tokens is a generic representation to use as an input -- many different types of data can be “tokenised'' and transformers are then immediately applicable rather than requiring a bespoke architectures for each modality as was previously the case (CNNs for images, RNNs for sequences, deepsets for sets etc.). Moreover, this means that you don’t need bespoke handcrafted architectures for mixing data of different modalities --- you can just throw them all into a big set of tokens.

% \begin{equation}
% \mathcal{R}^{\text{d}} = g^e_{\sigma_{2}}\left(\frac{\Gamma^z}{Q_{12}^2-M_W^2}\right)
% \end{equation}
% Proin convallis interdum libero a sollicitudin. In dignissim quam id viverra congue. Pellentesque eget magna massa. Quisque sit amet sagittis felis. Proin a ipsum quis magna sodales egestas non euismod turpis. Sed nisi purus, vestibulum vitae volutpat auctor, euismod sit amet mauris. 

% \begin{margintable}
% \centering
% \begin{tabular}{||c c c c||} 
%  \hline
%  Col1 & Col2 & Col2 & Col3 \\ [0.5ex] 
%  \hline\hline
%  1 & 6 & 87837 & 787 \\ 
%  2 & 7 & 78 & 5415 \\
%  3 & 545 & 778 & 7507 \\
%  4 & 545 & 18744 & 7560 \\
%  5 & 88 & 788 & 6344 \\ [1ex] 
%  \hline
% \end{tabular}
% \captionsetup{justification=centering}
% \caption{Table to test captions and labels}
% \end{margintable}

\subsection{Goal: representations of sequences}

The transformer will ingest the input data $X^{(0)}$ and return a representation of the sequence in terms of another matrix $X^{(M)}$ which is also of size $D \times N$.  The slice $\bm{x}_n = X^{(M)}_{:,n}$ will be a vector of features representing the sequence at the location of token $n$. These representations can be used for auto-regressive prediction of the next (n+1)th token, global classification of the entire sequence (by pooling across the whole representation), sequence-to-sequence or image-to-image prediction problems, etc. Here $M$ denotes the number of layers in the transformer.

\section{The transformer block}

The representation of the input sequence will be produced by iteratively applying a transformer block \[X^{(m)} = \text{transformer-block}(X^{(m-1)}).\] The block itself comprises two stages: one operating across the sequence and one operating across the features. The first stage refines each feature independently according to relationships between tokens across the sequence e.g.~how much a word in a sequence at position $n$ depends on previous words at position $n'$, or how much two different patches from an image are related to one another. This stage acts horizontally across rows of $X^{(m-1)}$.  The second stage refines the features representing each token. This stage acts vertically across a column of $X^{(m-1)}$. By repeatedly applying the transformer block the representation at token $n$ and feature $d$ can be shaped by information at token $n'$ and feature $d'$.\sidenote{\footnotesize The idea of interleaving processing across the sequence and across features is a common motif of many machine learning architectures including graph neural networks (interleaves processing across nodes and across features), Fourier neural operators (interleaves processing across space and across features), and bottleneck blocks in ResNets (interleaves processing across pixels and across features).}

\subsection{Stage 1: self-attention across the sequence}
The output of the first stage of the transformer block is another $D \times N$ array, $Y^{(m)}$. The output is produced by aggregating information across the sequence independently for each feature using an operation called \emph{attention}. 

\textbf{Attention.} Specifically, the output vector at location $n$, denoted $\bm{y}^{(m)}_n$, is produced by a simple weighted average of the input features at location $n'=1 \hdots N$, denoted $\bm{x}^{(m-1)}_{n'}$, that is\sidenote{\footnotesize\textbf{Relationship to Convolutional Neural Networks} (CNNs). The attention mechanism can recover convolutional filtering as a special case e.g.~if $\bm{x}^{(0)}_{n}$ is a 1D regularly sampled time-series and $ A^{(m)}_{n',n} =  A^{(m)}_{n'-n}$ then the attention mechanism in eq.~\ref{eq:attention} becomes a convolution. Unlike normal CNNs, these filters have full temporal support. Later we will see that the filters themselves dynamically depend on the input, another difference from standard CNNs. We will also see a similarity: transformers will use multiple attention maps in each layer in the same way that CNNs use multiple filters (though typically transformers have fewer attention maps than CNNs have channels).}
\begin{equation} \bm{y}^{(m)}_n = \sum_{n'=1}^N \bm{x}^{(m-1)}_{n'} A^{(m)}_{n',n}. \label{eq:attention} \end{equation}

Here the weighting is given by a so-called \emph{attention matrix} $A^{(m)}_{n',n}$ which is of size\sidenote{\footnotesize The need for transformers to store and compute $N \times N$ attention arrays can be a major computational bottleneck, which makes processing of long sequences challenging.} $N \times N$ and normalises over its columns $\sum_{n'=1}^N A^{(m)}_{n',n}=1$. Intuitively speaking $A^{(m)}_{n',n}$ will take a high value for locations in the sequence $n'$ which are of high relevance for location $n$. For irrelevant locations, it will take the value $0$. For example, all patches of a visual scene coming from a single object might have high corresponding attention values. 

We can compactly write the relationship as a matrix multiplication, 
\begin{align}
Y^{(m)} = X^{(m-1)} A^{(m)} \label{eq:SA1},
\end{align}
and we illustrate it below in figure \ref{fig:attention}.\sidenote{\footnotesize When training transformers to perform auto-regressive prediction, e.g.~predicting the next word in a sequence based on the previous ones, a clever modification to the model can be used to accelerate training and inference. This involves applying the transformer to the whole sequence, and using masking in the attention mechanism ($A^{(m)}$ becomes an upper triangular matrix) to prevent future tokens affecting the representation at earlier tokens. Causal predictions can then be made for the entire sequence in one forward pass through the transformer. See section \ref{sec:head} for more information.}   
% figure out of margin
\begin{figure}[!h]\centering 
 \includegraphics[scale = 0.35]{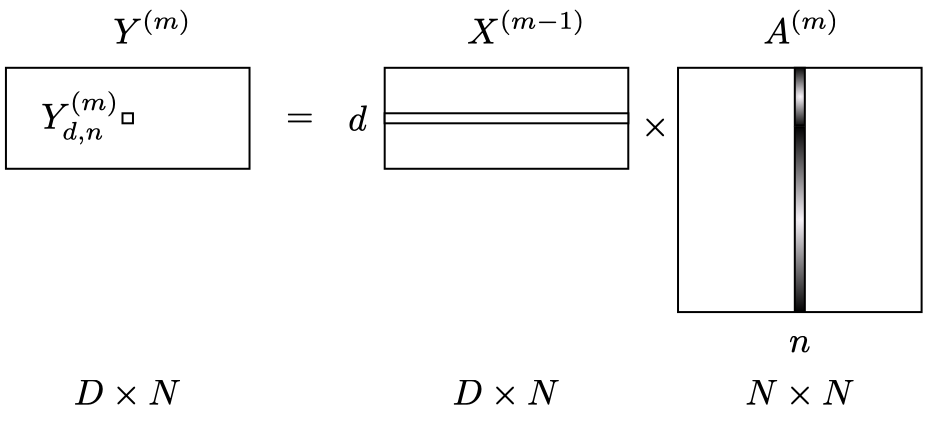}
 \caption{ The output of an element of the attention mechanism, $Y_{d,n}^{(m)}$, is produced by the dot product of the input horizontally sliced through time $X_{d,:}^{(m)}$ with a vertical slice from the attention matrix $A^{(m)}_{:,n}$. Here the shading in the attention matrix represent the elements with a high value in white and those with a low value, near to 0, in black.}
 \label{fig:attention}
 \end{figure}

\textbf{Self-attention.} So far, so simple. But where does the attention matrix come from? The neat idea in the first stage of the transformer is that the attention matrix is generated from the input sequence itself -- so-called \emph{self-attention}. 

A simple way of generating the attention matrix from the input would be to measure the similarity between two locations by the dot product between the features at those two locations and then use a softmax function to handle the normalisation i.e.\sidenote{\footnotesize We temporarily suppress the superscripts here to ease the notation so $A^{(m)}_{n,n'}$ becomes $A_{n,n'}$ and similarly $\bm{x}^{(m)}_{n}$ becomes $\bm{x}_{n}$. }
\[ A_{n,n'} = \frac{\exp( \bm{x}_n^\top  \bm{x}_{n'})}{\sum_{n''=1}^N \exp(\bm{x}_{n''}^\top \bm{x}_{n'})}. \]
However, this na\"ive approach entangles information about the similarity between locations in the sequence with the content of the sequence itself. 

An alternative is to perform the same operation on a linear transformation of the sequence, $U \bm{x}_n$, so that\sidenote{\footnotesize Often you will see attention parameterised as \[ A_{n,n'} = \frac{\exp( \bm{x}_n^\top U^\top U \bm{x}_{n'} /\sqrt{D})}{\sum_{n''=1}^N \exp(\bm{x}_{n''}^\top U^\top U \bm{x}_{n'}/\sqrt{D})}. \]  Dividing the exponents by the square-root of the dimensionality of the projected vector helps numerical stability, but in this presentation we absorb this term into $U$ to improve clarity. }
\[ A_{n,n'} = \frac{\exp( \bm{x}_n^\top U^\top U \bm{x}_{n'})}{\sum_{n''=1}^N \exp(\bm{x}_{n''}^\top U^\top U \bm{x}_{n'})} \]
Typically, $U$ will project to a lower dimensional space i.e. $U$ is $K \times D$ dimensional with $K<D$. In this way only some of the features in the input sequence need be used to compute the similarity, the others being projected out, thereby decoupling the attention computation from the content. However, the numerator in this construction is symmetric. 
This could be a disadvantage. For example, we might want the word `caulking iron' to be strongly associated with the word `tool' (as it is a type of tool), but have the word `tool' more weakly associated with the word `caulking iron' (because most of us rarely encounter it).\sidenote{\footnotesize Some of this effect could be handled by the normalisation in the denominator, but asymmetric similarity allows more flexibility. However, I do not know of experimental evidence to support using $U_{\bm{q}} \ne U_{\bm{k}}$.}

Fortunately, it is simple to generalise the attention mechanism above to be asymmetric by applying two different linear transformations to the original sequence,
\begin{equation} 
A_{n,n'} = \frac{\exp \left( \bm{x}_n^\top U_{\bm{k}}^\top U_{\bm{q}}^{} \bm{x}_{n'} \right)}{\sum_{n''=1}^N \exp \left(\bm{x}_{n''}^\top U_{\bm{k}}^\top U_{\bm{q}}^{} \bm{x}_{n'}\right)}. \label{eq:SA2}
\end{equation}
The two quantities that are dot-producted together here $\bm{q}_n = U_{\bm{q}}\bm{x}_{n}$ and $\bm{k}_n = U_{\bm{k}} \bm{x}_{n}$ are typically known as the \emph{queries} and the \emph{keys}, respectively.

Together equations \ref{eq:SA1} and \ref{eq:SA2} define the self-attention mechanism. Notice that the $K \times D$ matrices $U_{\bm{q}}$ and $U_{\bm{k}}$ are the only parameters of this mechanism.\sidenote{\footnotesize\textbf{Relationship to Recurrent Neural Networks} (RNNs). It is illuminating to compare the temporal processing in the transformer to that of RNNs which recursively update a hidden state feature representation ($\bm{x}^{(1)}_n$) based on the current observation ($\bm{x}^{(0)}_n$) and the previous hidden state  $\bm{x}^{(1)}_n = f(\bm{x}^{(1)}_{n-1}; \bm{x}^{(0)}_n) = f(f(\bm{x}^{(1)}_{n-2}; \bm{x}^{(0)}_{n-1}); \bm{x}^{(0)}_n)$. Here we've unrolled the RNN one step to show that observations which are nearby to the hidden state (e.g.~$\bm{x}^{(0)}_n$) are treated differently from observations that are further away (e.g.~$\bm{x}^{(0)}_{n-1}$), as information is propagated by recurrent application of the function $f(\cdot)$. In contrast, in the transformer, self-attention treats all observations at all time-points in an identical manner, no matter how far away they are. This is one reason why they find it simpler to learn long-range relationships.} 

\textbf{Multi-head self-attention (MHSA).} In the self-attention mechanisms described above, there is one attention matrix which describes the similarity of two locations within the sequence. This can act as a bottleneck in the architecture -- it would be useful for pairs of points to be similar in some `dimensions' and different in others.\sidenote{\footnotesize If attention matrices are viewed as a data-driven version of filters in a CNN, then the need for more filters / channels is clear. Typical choices for the number of heads $H$ is 8 or 16, lower than typical numbers of channels in a CNN.} 

In order to increase capacity of the first self-attention stage, the transformer block applies $H$ sets of self-attention in parallel\sidenote{\footnotesize The computational cost of multi-head self-attention is usually dominated by the matrix multiplication involving the attention matrix and is therefore $\mathcal{O}(H D N^2)$.} (termed $H$ heads) and then linearly projects the results down to the $D \times N$ array required for further processing. This slight generalisation is called \emph{multi-head self-attention}. 
\begin{align}
Y^{(m)} = \text{MHSA}_{\theta}(X^{(m-1)}) = \sum_{h=1}^H V^{(m)}_h X^{(m-1)} A_h^{(m)}, \label{eq:MHSA}\;\;\text{where} \;\;\\
%
%
%
%[A^{(m)}_{h}]_{n,n'} = \frac{\exp\left( \left(\bm{x}_n^{(m-1)}\right)^\top \left(U_{h,q}^{(m)} \right)^\top U^{(m)}_{h,k} \bm{x}^{(m-1)}_{n'} \right)}{\sum_{n''=1}^N \exp\left( \left(\bm{x}^{(m-1)}_{n''}\right)^\top \left(U^{(m)}_{h,q}\right)^\top U^{(m)}_{h,v} \bm{x}_{n'} \right)} \notag. 
[A^{(m)}_{h}]_{n,n'} = \frac{\exp\left( \left(\bm{k}_{h,n}^{(m)}\right)^\top  \bm{q}_{h,n'}^{(m)} \right)}{\sum_{n''=1}^N \exp \left( \left(\bm{k}_{h,n''}^{(m)}\right)^\top   \bm{q}_{h,n'}^{(m)} \right)} \\ \bm{q}_{h,n}^{(m)} = U^{(m)}_{\bm{q},h} \bm{x}^{(m-1)}_{n} \;\; \text{and} \;\;\bm{k}_{h,n}^{(m)} = U^{(m)}_{\bm{k},h} \bm{x}^{(m-1)}_{n}.
\end{align}
Here the $H$ matrices $V^{(m)}_h$ which are $D \times D$ project the $H$ self-attention stages down to the required output dimensionality $D$.\sidenote{\footnotesize The product of the matrices $V^{(m)}_h X^{(m-1)}$ is related to the so-called \emph{values} which are normally introduced in descriptions of self-attention along side queries and keys. In the usual presentation, there is a redundancy between the linear transform used to compute the values and the linear projection at the end of the multi-head self-attention, so we have not explicitly introduced them here. The standard presentation can be recovered by setting $V_h$ to be a low-rank matrix $V_h = U_h U_{\bm{v},h}$ where $U_h$ is $D$x$K$ and $U_{\bm{v},h}$ is $K$x$D$. Typically $K$ is set to $K = D/H$ so that changing the number of heads leads to models with similar numbers of parameters and computational demands. } 

The addition of the matrices $V^{(m)}_h$, and the fact that retaining just the diagonal elements of the attention matrix $A^{(m)}$ will interact the signal instantaneously with itself, does mean there is some cross-feature processing in multi-head self-attention, as opposed to it containing purely cross-sequence processing. However, the stage has limited capacity for this type of processing and it is the job of the second stage to address this.

\begin{figure*}[!htbp]\centering\footnotesize
 \includegraphics[scale = 0.3]{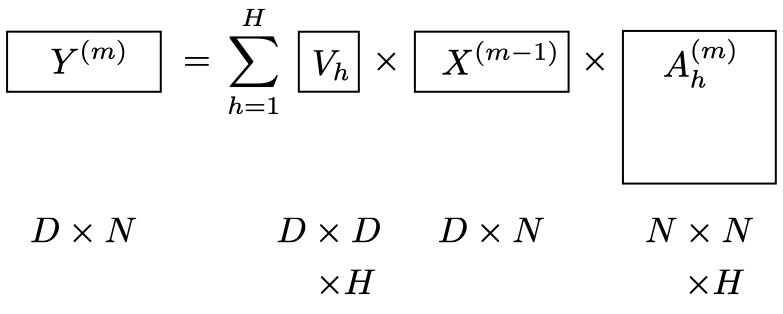}
 \caption{\footnotesize Multi-head self-attention applies $H$ self-attention operations in parallel and then linearly projects the $HD \times N$ dimensional output down to $D \times N$ by applying a linear transform, implemented here by the $H$ matrices $V_h$. }
 \label{fig:mhsa}
 \end{figure*}

Figure \ref{fig:mhsa} shows multi-head self-attention schematically. Multi-head attention comprises the following parameters $\theta = \{ U_{\bm{q},h},U_{\bm{k},h},V_{h} \}_{h=1}^H$ i.e.~$3H$ matrices of size $K \times D$, $K \times D$, and $D \times D$ respectively.
%
% figure out of margin
%

%\[ Y^{(m)} = X^{(m)} A^{(m)} = \text{SA}_{\theta}(X^{(m-1)})\]
%\[ A^{(m)}_{n,n'} = \frac{\exp( \mathbf{q}_n^\top \bm{k}_{n'})}{\sum_{n''=1}^N \exp(\bm{q}_{n''}^\top \bm{k}_{n'})}\]

\subsection{Stage 2: multi-layer perceptron across features}

The second stage of processing in the transformer block operates across features, refining the representation using a non-linear transform. To do this, we simply apply a multi-layer perceptron (MLP) to the vector of features at each location $n$ in the sequence, 
\[ \bm{x}_n^{(m)} = \text{MLP}_{\theta}(\bm{y}_n^{(m)}).\]
Notice that the parameters of the MLP, $\theta$, are the same for each location $n$.\sidenote{\footnotesize The MLPs used typically have one or two hidden-layers with dimension equal to the number of features $D$ (or larger). The computational cost of this step is therefore roughly $N \times D \times D$. If the feature embedding size approaches the length of the sequence $D \approx N$, the MLPs can start to dominate the computational complexity (e.g.~this can be the case for vision transformers which embed large patches).} \sidenote{\footnotesize \textbf{Relationship to Graph Neural Networks (GNNs)}. At a high level, graph neural networks interleave two steps. First, a message passing step where each node receives messages from its neighbours which are then aggregated together. Second, a feature processing step where the incoming aggregated messages are used to update each node's features. Through this lens, the transformer can be viewed as an unrolled GNN with each token corresponding to an edge of a fully connected graph. MHSA forms the message passing step, and the MLPs forming the feature update step. Each transformer block corresponds to one update of the GNN. Moreover, many methods for scaling transformers introduce sparse forms of attention where each token  attends to only a restricted set of other tokens, that is they specify a sparse graph connectivity structure. Arguably, in this way transformers are more general as they can use different graphs at different layers in the transformer.}

\subsection{The transformer block: Putting it all together with residual connections and layer normalisation}

We can now stack MHSA and MLP layers to produce the transformer block. Rather than doing this directly, we make use of two ubiquitous transformations to produce a more stable model that trains more easily: residual connections and normalisation.

\textbf{Residual connections.} The use of residual connections is widespread across machine learning as they make initialisation simple, have a sensible inductive bias towards simple functions, and stabilise learning \citep{resnets}. Instead of directly specifying a function $x^{(m)} = f_{\theta}(x^{(m-1)})$, the idea is to parameterise it in terms of an identity mapping and a residual term 
\[ x^{(m)} = x^{(m-1)} + \text{res}_{\theta}(x^{(m-1)}).\] 
Equivalently, this can be viewed as modelling the differences  between the representation $x^{(m)} - x^{(m-1)} = \text{res}_{\theta}(x^{(m-1)})$ and will work well when the function that is being modelled is close to identity. This type of parameterisation is used for both the MHSA and MLP stages in the transformer, with the idea that each applies a mild non-linear transformation to the representation. Over many layers, these mild non-linear transformations compose to form large transformations.

%
%\begin{marginfigure}%
\begin{figure*}[!htbp]\centering
   \includegraphics[width=0.90\linewidth]{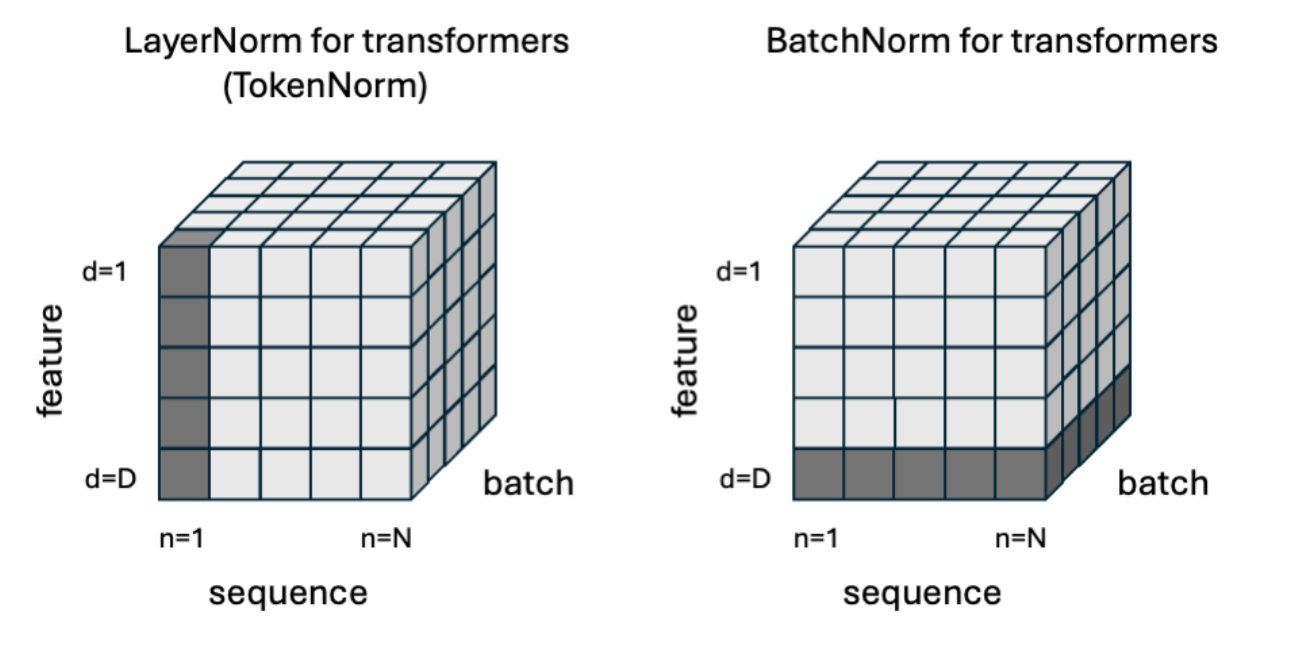}
   \caption{\footnotesize Transformers perform layer normalisation (left hand schematic) which normalises the mean and standard deviation of each individual token in each sequence in the batch. Batch normalisation (right hand schematic), which normalises over the sequence \emph{and} batch dimension together, is found to be far less stable \citep{pmlr-v119-shen20e}.}
   \label{fig:trans-norm}
   \end{figure*}
 %\end{marginfigure}

\textbf{Token normalisation.} The use of normalisation, such as LayerNorm and BatchNorm, is also widespread across the deep learning community as a means to stabilise learning. There are many potential choices for how to compute normalisation statistics (see figure  \ref{fig:trans-norm} for a discussion), but the standard approach is to use LayerNorm \citep{ba2016layernorm} which normalises each token separately, removing the mean and dividing by the standard deviation,\sidenote{\footnotesize This is also known as z-scoring in some fields and is related to whitening.}  
\[ \bar{x}_{d,n}  = \frac{1}{\sqrt{\text{var}(\bm{x}_n)}} \left( x_{d,n} - \text{mean}(\bm{x}_n) \right) \gamma_d + \beta_d = \text{LayerNorm}(X)_{d,n} \] 
where $ \text{mean}(\bm{x}_n) = \frac{1}{D} \sum_{d=1}^D x_{d,n} $ and $\text{var}(\bm{x}_n) = \frac{1}{D} \sum_{d=1}^D (x_{d,n} - \text{mean}(\bm{x}_n))^2.$ The two parameters $\gamma_d$ and $\beta_d$ are a learned scale and shift. 

\begin{marginfigure}%
   \includegraphics[width=0.95\linewidth]{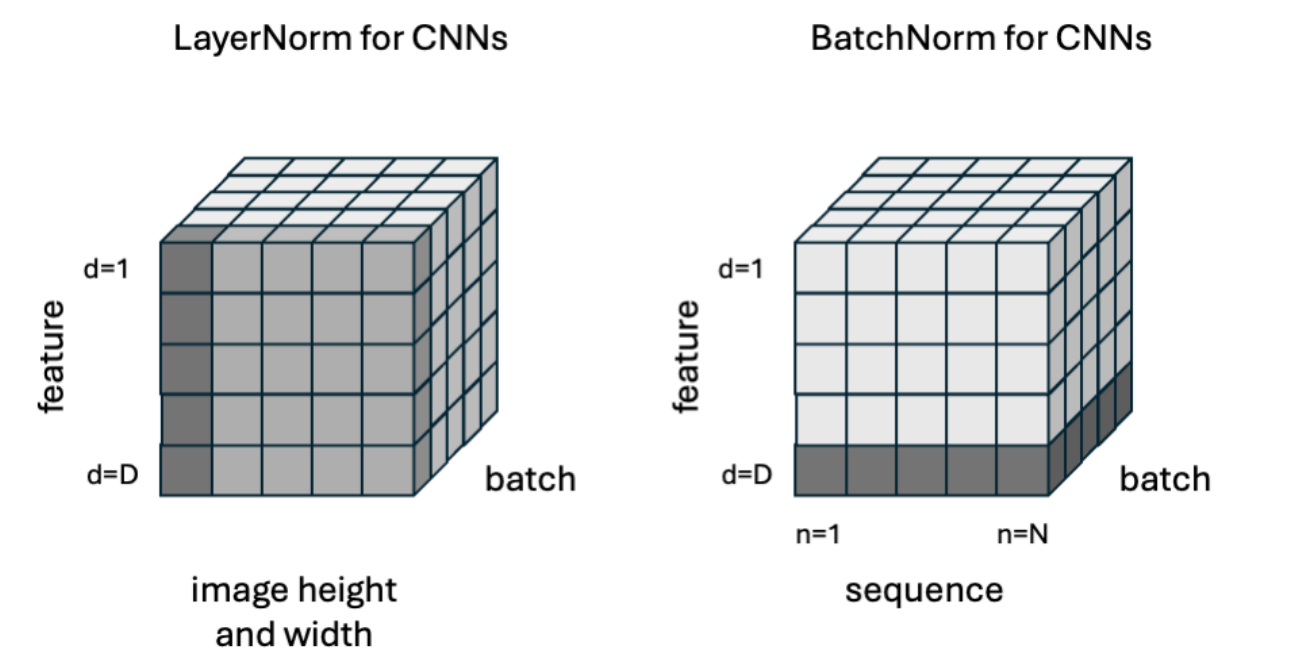}
   \caption{In CNNs LayerNorm is ambiguous  applied sometimes referring to normalising across both the features \emph{and} the feature maps (i.e.~across the height and width of the images) and sometimes just across the features (left hand schematic). As the height and width dimension in CNNs corresponds to the sequence dimension, $1\hdots N$ of transformers, the term 'LayerNorm' is arguably used inconsistently (compare to figure \ref{fig:trans-norm}). I would prefer to call the normalisation used in transformers 'token normalisation' instead to avoid confusion. Batch normalisation (right hand schematic) is consistently defined.}
   \label{fig:cnn-norm}
 \end{marginfigure}

As this transform normalises each token individually and as LayerNorm is sometimes applied differently in CNNs, see figure \ref{fig:cnn-norm}, I would prefer to call this normalisation TokenNorm. 

This transform stops feature representations blowing up in magnitude as non-linearities are repeatedly applied through neural networks.\sidenote{\footnotesize Whilst it is possible to control the non-linearities and weights in neural networks to prevent explosion of the representation, the constraints this places on the activation functions can adversely affect learning. The LayerNorm approach is arguably simpler and simpler to train.} In transformers, LayerNorm is usually applied in the residual terms of both the MHSA and MLP stages.

\begin{figure}[!h]\centering
 \includegraphics[scale = 0.24]{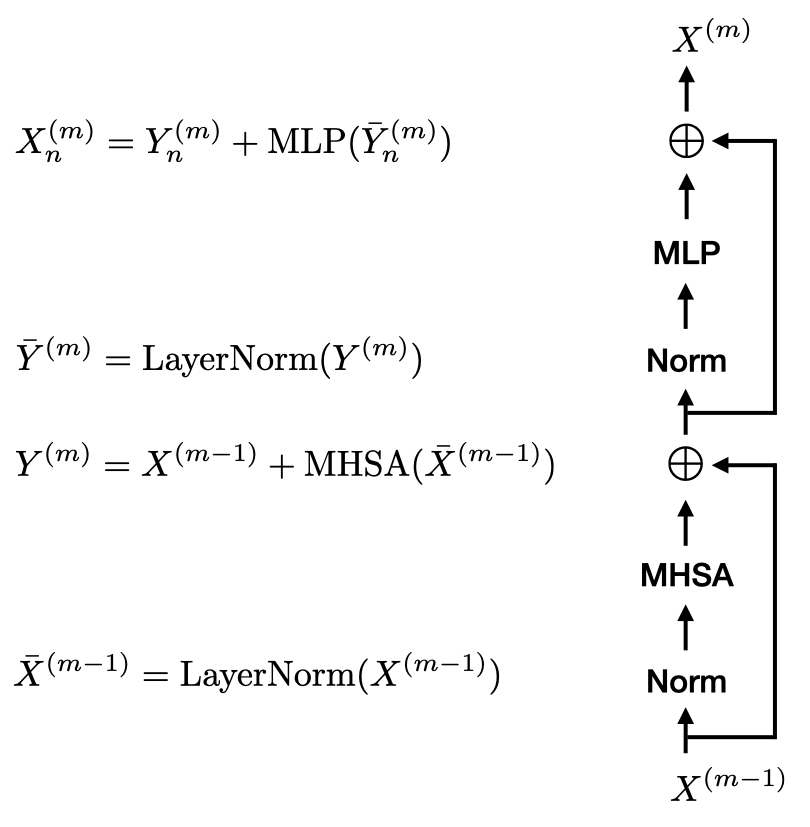}
 \caption{The transformer block. Residual connections are added to the multi-head self-attention (MHSA) stage and the multi-layer perceptron (MLP) stage. Layer normalisation is also applied to the inputs of both the MHSA and the MLP. They are then stacked. This block can then be repeated $M$ times.}
 \label{fig:transformer-block}
 \end{figure}

Putting this all together, we have the standard transformer block shown schematically in figure \ref{fig:transformer-block}.\sidenote{\footnotesize The exact configuration of the normalisation and residual layers can differ, but here we show a standard setup \citep{pmlr-v119-xiong20b}.} 

\section{Position encoding} \label{sec:pos-ecoding}

The transformer treats the data as a set --- if you permute the columns of $X^{(0)}$ (i.e.~re-order the tokens in the input sequence) you permute all the representations throughout the network $X^{(m)}$ in the same way. This is key for many applications since there may not be a natural way to order the original data into a sequence of tokens. For example, there is no single ‘correct’ order to map image patches into a one dimensional sequence. 

However, this presents a problem since positional information is key in many problems and the transformer has thrown it out. The sequence ‘herbivores eat plants’ should not have the same representation (up to permutation) as ‘plants eat herbivores’. Nor should an image have the same representation as one comprising the same patches randomly permuted. Thankfully, there is a simple fix for this: the location of each token within the original dataset should be included in the token itself, or through the way it is processed. There are several options how to do this, one is to include this information directly into the embedding $X^{(0)}$. E.g.~by simply adding the position embedding (surprisingly this works\sidenote{\footnotesize Vision transformers \citep{vision-transformer} use $\bm{x}_n^{(0)} = W \bm{p}_n + \bm{e}_n$ where $\bm{p}_n$ is the $n$th vectorised patch, $\bm{e}_n$ is the learned position embedding, and $W$ is the patch embedding matrix. Arguably it would be more intuitive to append the position embedding to the patch embedding. However, if we use the concatenation approach and consider what happens after applying a linear transform,
\begin{align}  V \left[\begin{array}{c} W \bm{p}_n \\ \bm{e}_n \end{array} \right] & =  \left[\begin{array}{cc} V_{11} & V_{12} \\ V_{21} & V_{22} \end{array} \right] \left[\begin{array}{c} W \bm{p}_n \\ \bm{e}_n \end{array} \right] \nonumber \\
& = 
\left[\begin{array}{c} V_{11} W \bm{p}_n + V_{12} \bm{e}_n \\
V_{21} W \bm{p}_n + V_{22} \bm{e}_n
\end{array} \right] \nonumber\\
& = W' \bm{p}_n + \bm{e}_n' \nonumber
\end{align}
we recover the additive construction, which is one hint as to why the additive construction works.}) or concatenating. The position information can be fixed e.g.~adding a vector of sinusoids of different frequencies and phases to encode position of a word in a sentence \citep{NIPS2017_3f5ee243}, or it can be a free parameter which is learned \citep{devlin-etal-2019-bert}, as it often done in image transformers. There are also approaches to include relative distance information between pairs of tokens by modifying the self-attention mechanism \citep{9710917} which connects to equivariant transformers.

\section{Application specific transformer variants} \label{sec:head}
For completeness we will give some simple examples for how the standard transformer architecture above is used and modified for specific applications. This includes adding a head to the transformer blocks to carry out the desired prediction task, but also modifications to the standard construction of the body.

\subsection{Auto-regressive language modelling}

In auto-regressive language modelling  the goal is to predict the next word $w_n$ in the sequence given the previous words $w_{1:n-1}$, that is to return $p(w_n = w | w_{1:n-1})$. Two modifications are required to use the transformer for this task --- a change to the body to make the architecture efficient and the addition of a head to make the predictions for the next word.

\textbf{Modification to the body: auto-regressive masking.} Applying the version of the transformer we have covered so far to auto-regressive prediction is computationally expensive, both during training and testing. To see this, note that AR prediction requires making a sequence of predictions: you start by predicting the first word $p(w_1 = w)$, then you predict the second given the first $ p(w_2 = w | w_{1})$, then the third word given the first two $ p(w_2 = w | w_{1},w_2)$, and so on until you predict the last item in the sequence $p(w_N = w | w_{1:N-1})$. This requires applying the transformer $N-1$ times with input sequences that grow by one word each time: ${w_1,w_{1:2},\hdots,w_{1:N-1}}$. This is very costly at both training-time and test-time. 

Fortunately, there is a neat way around this by enabling the transformer to support incremental updates whereby if you add a new token to an existing sequence, you do not change the representation for the old tokens. To make this property clear, I will define it mathematically: let the output of the incremental transformer applied to the first $n$ words be denoted\sidenote{\footnotesize Note that I'm overloading the notation here: previously superscripts denoted layers in the transformer, but here I'm using them to denote the number of items in the input sequence.}
\[
X^{(n)} = \text{transformer-incremental}(w_{1:n}).
\]
Then the output of the incremental transformer when applied to $n+1$ words is
\[
X^{(n+1)} = \text{transformer-incremental}(w_{1:n+1}).
\]
In the incremental transformer $X^{(n)} = X^{(n+1)}_{1:D,1:n}$ i.e.~the representation of the old tokens has not changed by adding the new one. If we have this property then 1.~at test-time auto-regressive generation can use incremental updates to compute the new representation efficiently, 2.~at training time we can make the $N$ auto-regressive predictions for the whole sequence $p(w_1 = w) p(w_2 = w | w_{1}) p(w_2 = w | w_{1},w_2) \hdots p(w_N = w | w_{1:N-1})$ in a single forwards pass.

Unfortunately, the standard transformer introduced above does not have this property due to the form of the attention used. Every token attends to every other token,  so if we add a new token to the sequence then the representation for every token changes throughout the transformer. However, if we mask the attention matrix so that it is upper-triangular $A_{n,n'} = 0$ when $n>n'$ then the representation of each word \emph{only depends on the previous words}.\sidenote{\footnotesize Notice that this masking operation also encodes position information since you can infer the order of the tokens from the mask.} This then gives us the incremental property as none of the other operations in the transformer operate across the sequence.\sidenote{\footnotesize This restriction to the attention will cause a loss of representational power. It's an open question as to how significant this is and whether increasing the capacity of the model can mitigate it e.g.~by using higher dimensional tokens, i.e.~increasing $D$.}

\textbf{Adding a head.} We're now almost set to perform auto-regressive language modelling. We apply the masked transformer block $M$ times to the input sequence of words. We then take the representation at token $n-1$, that is $\bm{x}^{(M)}_{n-1}$ which captures causal information in the sequence at this point, and 
generate the probability of the next word through a softmax operation  
\[ p(w_n = w | w_{1:n-1}) = p(w_n = w | \bm{x}^{(M)}_{n-1}) = \frac{ \exp(\bm{g}_w^\top \bm{x}_{n-1}^{(M)}) } {\sum_{w=1}^W\exp(\bm{g}_w^\top \bm{x}_{n-1}^{(M)})}.\]
Here $W$ is the vocabulary size, the wth word is $w$ and $\{ \bm{g}_w \}_{w=1}^W$ are softmax weights that will be learned.

\subsection{Image classification}

For image classification the goal is to predict the label $y$ given the input image which has been tokenised into the sequence $X^{(0)}$, that is $p(y | X^{(0)} )$. One way of computing this distribution  would be to apply the standard transformer body $M$ times to the tokenised image patches before aggregating the final layer of the transformer, $X^{(M)}$, across the sequence e.g.~by spatial pooling $\bm{h} = \sum_{n=1}^N \bm{x}_n^{(M)}$ in order to form a feature representation for the entire image. The representation $\bm{h}$ could then be used to perform softmax classification. An alternative approach is found to perform better \citep{vision-transformer}. Instead we introduce a new fixed (learned) token at the start $n=0$ of the input sequence $\bm{x}_0^{(0)}$. At the head we use the $n=0$ vector, $\bm{x}_0^{(M)}$, to perform the softmax classification. This approach has the advantage that the transformer maintains and refines a global representation of the sequence at each layer $m$ of the transformer that is appropriate for classification.  

\subsection{More complex uses}

The transformer block can also be used as part of more complicated systems e.g.~in encoder-decoder architectures for sequence-to-sequence modelling for translation \citep{devlin-etal-2019-bert,NIPS2017_3f5ee243} or in masked auto-encoders for self-supervised vision systems \citep{he2021masked-autoencoders}.

\section{Conclusion}

This concludes this basic introduction to transformers which aspired to be mathematically precise and to provide intuitions behind the design decisions. 

We have not talked about loss functions or training in any detail, but this is because rather standard deep learning approaches are used for these. Briefly, transformers are typically trained using the Adam optimiser. They are often slow to train compared to other architectures and typically get more unstable as training progresses. Gradient clipping, decaying learning rate schedules, and increasing batch sizes through training help to mitigate these instabilities, but often they still persist.

\vspace{3mm}
{\bf Acknowledgements.} We thank Dr.~Max Patacchiola, Sasha Shysheya, John Bronskill, Runa Eschenhagen and Jess Riedel for feedback on previous versions of this note. Richard E.~Turner is supported by Microsoft, Google, Amazon, ARM, Improbable and EPSRC grant EP/T005386/1.
%\normalsize

\bibliographystyle{plainnat}
\bibliography{bibliography}

\end{document}